\documentclass[lettersize,journal]{IEEEtran}

\usepackage{amsmath,amsfonts}
\usepackage{algorithm}
\usepackage{array}
\usepackage[caption=false,font=normalsize,labelfont=sf,textfont=sf]{subfig}
\usepackage{textcomp}
\usepackage{stfloats}
\usepackage{url}
\usepackage{verbatim}
\usepackage{graphicx}
\usepackage{cite}
\usepackage{hhline}
\usepackage{amssymb}
\usepackage{booktabs}
\usepackage{siunitx}
\usepackage{xcolor}
\usepackage{multirow}
\usepackage{makecell}
\usepackage{times}
\usepackage{epsfig}
\usepackage{hyperref}
\usepackage{colortbl}
      
\graphicspath{{./fig/}}
\usepackage[noend]{algpseudocode}
\usepackage{diagbox}
\usepackage{booktabs}

\newcommand{\myPara}[1]{\vspace{.05in}\noindent\textbf{#1}}

\begin{document}

\title{Latent Spatiotemporal Adaptation for Generalized Face Forgery Video Detection}

\author{Daichi Zhang,
        Zihao Xiao,
        Jianmin Li,
        and Shiming~Ge,~\IEEEmembership{Senior Member,~IEEE}% <-this % stops a space
\thanks{Daichi Zhang and Shiming Ge are with the Institute of Information Engineering, Chinese Academy of Sciences, Beijing 100092, China, and University of Chinese Academy of Sciences, Beijing 100049, China. Email: \{zhangdaichi, geshiming\}@iie.ac.cn.}% <-this % stops a space
\thanks{Jianmin Li is with the Department of Computer Science and Technology, Institute for AI, BNRist, Tsinghua University, Beijing 100084, China. E-mail: lijianmin@mail.tsinghua.edu.cn.}
\thanks{Zihao Xiao is with RealAI. E-mail: zihao.xiao@realai.ai.}
\thanks{Shiming Ge is the corresponding author. Email: geshiming@iie.ac.cn.}}% <-this %


%https://ieee-cas.org/publication/tcsvt/tcsvt-manuscript-submission-guide

\maketitle

\begin{abstract}
Face forgery videos have caused severe public concerns, and many detectors have been proposed. However, most of these detectors suffer from limited generalization when detecting videos from unknown distributions, such as from unseen forgery methods.
In this paper, we find that different forgery videos have distinct spatiotemporal patterns, which may be the key to generalization. 
To leverage this finding, we propose a Latent Spatiotemporal Adaptation~(LAST) approach to facilitate generalized face forgery video detection. The key idea is to optimize the detector adaptive to the spatiotemporal patterns of unknown videos in latent space to improve the generalization. 
Specifically, we first model the spatiotemporal patterns of face videos by incorporating a lightweight CNN to extract local spatial features of each frame and then cascading a vision transformer to learn the long-term spatiotemporal representations in latent space, which should contain more clues than in pixel space. Then by optimizing a transferable linear head to perform the usual forgery detection task on known videos and recover the spatiotemporal clues of unknown target videos in a semi-supervised manner, our detector could flexibly adapt to unknown videos' spatiotemporal patterns, leading to improved generalization. Additionally, to eliminate the influence of specific forgery videos, we pre-train our CNN and transformer only on real videos with two simple yet effective self-supervised tasks: reconstruction and contrastive learning in latent space and keep them frozen during fine-tuning.
Extensive experiments on public datasets demonstrate that our approach achieves state-of-the-art performance against other competitors with impressive generalization and robustness.

\end{abstract}

\begin{IEEEkeywords}
Face forgery video detection, spatiotemporal representation, domain adaptation, self-supervised learning.
\end{IEEEkeywords}

\section{Introduction}
\label{sec:intro}

\IEEEPARstart{F}{ace} forgery technology has been rapidly developed recently~\cite{DBLP:journals/corr/abs-1912-13457,thies2016face2face:}, especially after generative adversarial networks~(GANs) were proposed~\cite{goodfellow2014generative}. The generated videos can barely be distinguished by the human eye and can be easily produced by accessible online tools~\cite{deepfakes,Faceswap}. Potential attackers can easily take advantage of these techniques to mislead the public, defame celebrities or even fabricate evidence, leading to serious social, political and security consequences~\cite{DBLP:journals/tog/SuwajanakornSK17}. Thus, it is critical to develop effective face forgery video detectors.

\begin{figure}[t]
   \centering
   \includegraphics[width=\linewidth]{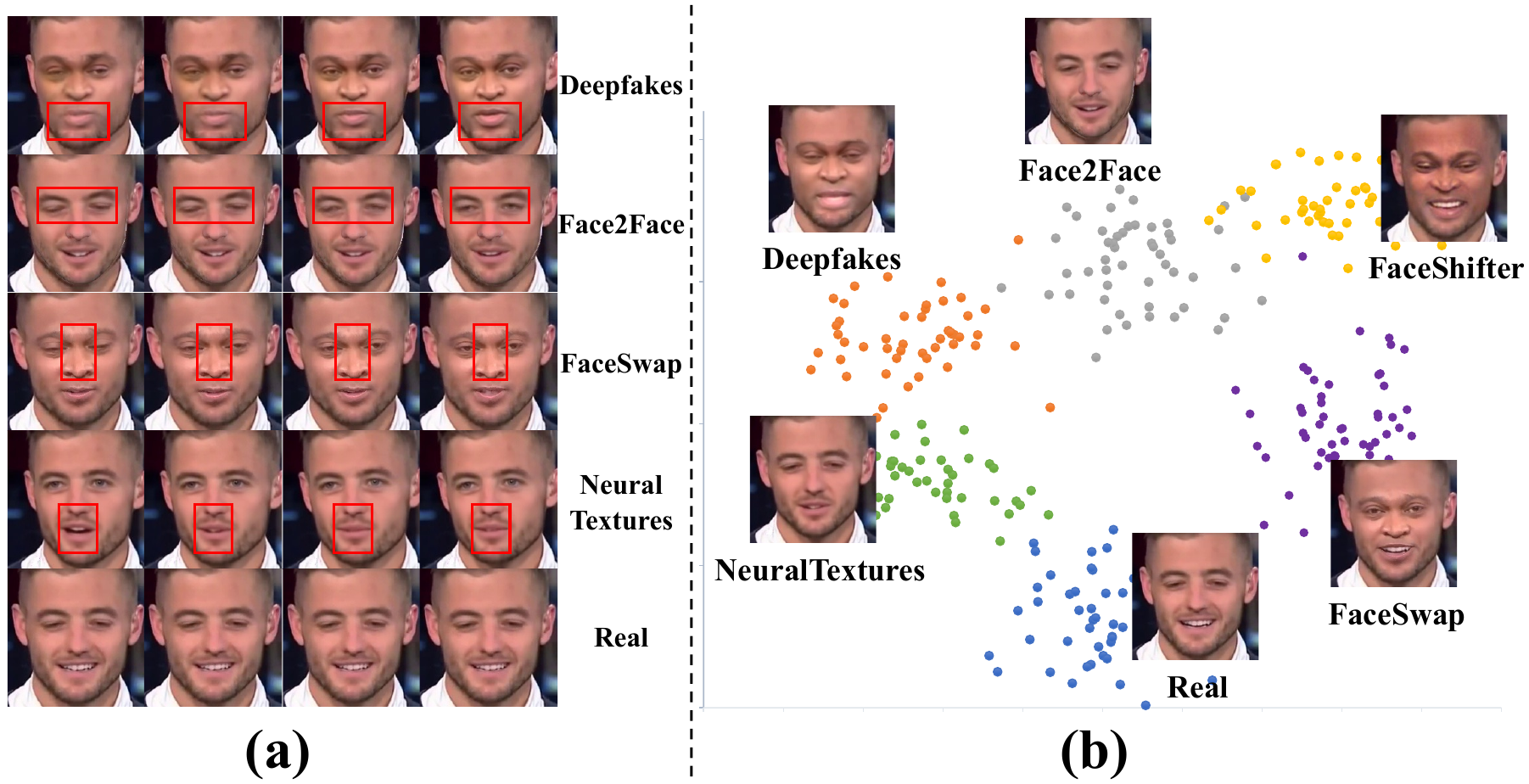}
   \caption{The distinct spatiotemporal patterns of real and different forgery videos (a) may cause the distribution gap in latent space when training a naive detector (b), leading to limited generalization.}
   \label{figure:mot}
   %\vspace{-.5em}
\end{figure}

Existing face forgery video detectors typically formulate the task as a binary classification task by exploring different features or patterns, such as mining the spatial or frequency pattern difference between real and fake images~\cite{Rssler2019FaceForensics,li2020face} or focusing on the spatiotemporal inconsistency existing in forgery videos~\cite{DBLP:conf/cvpr/SabirCJAMN19,DBLP:conf/ijcai/ZhangLL0G21,DBLP:journals/corr/abs-2207-06612,yu2023msvt}.
Though existing forgery video detectors achieve considerable performance~\cite{DBLP:conf/cvpr/LiL19c,DBLP:journals/corr/abs-2203-01265,DBLP:conf/ijcai/ZhangLL0G21,nguyen2019multi,DBLP:conf/wifs/AfcharNYE18,DBLP:conf/ijcai/HuXWLW021}, they may still suffer from limited generalization when detecting videos from unknown data distributions, such as from unseen forgery methods. 
In this paper, we find real and forgery videos generated by different forgery methods show distinct spatiotemporal patterns, as shown in Fig.~\ref{figure:mot} (a). These different spatiotemporal patterns in pixel space may cause the distribution gap in latent space when we train a naive detector\cite{Rssler2019FaceForensics}, as shown in Fig.~\ref{figure:mot} (b). Thus, if we train the detectors on specific forgery videos, they may suffer from the distribution gap caused by the spatiotemporal pattern difference, leading to limited generalization when detecting unknown videos. How to leverage the spatiotemporal difference may be the key to improve generalization.

To tackle this issue, there is one major insight in our mind: since it may be difficult to train a detector which is general enough to all different spatiotemporal patterns in unseen forgery videos with the forgery methods still evolving, but these unlabeled target videos are easily accessible in real scenarios, it may be more appropriate and practical to make the trained detector easily adaptive to the spatiotemporal patterns of unseen videos, thus leading to improved generalization. With the generative models still evolving, the forgery videos in the future are unpredictable and the detectors that are generalized well now may be unsuitable for future unknown methods. But if we have a detector that could be adaptive to any unknown videos without requiring their ground truth labels, it may be generalized and robust to the videos from any unknown sources or methods. 

Inspired by the above analysis, we propose the Latent Spatiotemporal Adaptation~(LAST) approach to achieve generalized face forgery video detection. Specifically, we first model the spatiotemporal patterns of face videos in latent space by incorporating a CNN to extract local spatial feature of each frame and then cascading a vision transformer to learn the long-term spatiotemporal representation, which should contain more useful clues compared to the redundant pixel space~\cite{DBLP:journals/corr/abs-2111-06377} and leverage the inherent advantages of CNN on local reception field~\cite{DBLP:conf/cvpr/HeZRS16} and transformer on long-term memory~\cite{DBLP:conf/iclr/DosovitskiyB0WZ21}. 
Then by optimizing a transferable linear head to perform the usual classification task and recover the spatiotemporal clues of unknown target videos in latent space in a semi-supervised manner, our detector could flexibly adapt to any unknown videos' spatiotemporal patterns, leading to improved generalization. This is also in the same spirit as multi-task learning that auxiliary tasks can help the main task learned better~\cite{argyriou2006multi}. Here, our specific auxiliary task is to recover the spatiotemporal clues in latent space.
To achieve this, we introduce another CNN reconstruction head to recover the local spatial features from the learned spatiotemporal representations of unknown videos, which could force the model learn more about the spatiotemporal clues of target unknown videos, thus mitigating their gap with known data distributions in latent space and improve the generalization. Note that during the adaptation, the ground truth labels of unknown target videos are not required, which could simulate the real scenarios for unknown or future generation methods.

Furthermore, if the initial spatiotemporal latent space is trained on specific forgery videos, the patterns from the specific forgery videos may affect the adaption process. Therefore, we further propose to pre-train our CNN and transformer only on real face videos with two simple yet effective self-supervised tasks and keep them frozen during adaption to learn the common spatiotemporal representation of face videos and fully eliminate the influence of specific forgery data. The two self-supervised tasks include reconstruction and contrastive learning in latent space. Specifically, reconstruction learning task is designed to reconstruct local spatial features from the learned spatiotemporal representation, which forces the representations retain more spatiotemporal clues of original face videos, and contrastive learning task enables the representations learned from the same video closer and different videos far away to explore both intra-video and inter-video connections. Both tasks should benefit learning the common spatiotemporal representation of face videos in latent space.

Our main contributions can be summarized as follows: 
\begin{itemize}
    \item We propose the Latent Spatiotemporal Adaptation~(LAST) approach, which first incorporates CNN for single-frame spatial feature cascaded with vision transformer for long-term spatiotemporal representation. By optimizing a linear head to perform the usual classification task and recover the spatiotemporal clues of unknown target videos in latent space, our method could flexibly adapt to unknown videos' spatiotemporal distribution while preserving discriminability. 
    \item We propose to pre-train our CNN and transformer on real-only face videos as initialization and keep them frozen during adaptation to leverage the common spatiotemporal representations of face videos with two simple yet effective self-supervised tasks: reconstruction and contrastive learning. This can effectively eliminate the effect of specific forgery videos in spatiotemporal latent space.
    \item Extensive experimental results demonstrate the superiority of our method compared to other state-of-the-art competitors with impressive generalization and robustness.
\end{itemize}

\section{Related Work}\label{sec:relate}

\subsection{Face Forgery Video Detection}
%Early face forgery video detectors focus on hand-crafted features to detect, such as~\cite{DBLP:conf/sswmc/0002SS07} leverages statistic feature, while recent many deep learning based detectors are proposed to detect face forgery videos by designed representation. 
Recent detectors mainly focus on different discriminative representations to detect, including using single frame features to train the detector, such as~\cite{Rssler2019FaceForensics} choose XceptionNet as backbone, ~\cite{li2020face} focuses on the blending boundary of forgery face, and~\cite{guo2023exposing,liu2023hierarchical} explore the frequency distribution to discriminate.
Some other detectors find spatiotemporal patterns hiding in forgery videos are important clues for detection, thus designing network to learn the spatiotemporal~\cite{DBLP:journals/corr/abs-2207-06612,DBLP:conf/aaai/GuCYDLM22,yu2023augmented} or temporal consistency~\cite{DBLP:conf/cvpr/SabirCJAMN19,DBLP:conf/iccv/ZhengB0ZW21,DBLP:conf/nips/GuanZHD0QZ22,yu2023msvt}. 
Although these methods have achieved considerable performance, they may still suffer from limited generalization when detecting unseen videos. 
%With the generation methods still evolving, how to achieve a generalized detector is still a critical issue now.
Therefore, one insight comes to our mind: since it may be difficult to train a detector that is generalized enough to detect all unseen forgery videos, it may be more appropriate and practical if we have a detector that can adapt to any target videos without requiring the ground truth labels. And the unlabeled target videos are easily accessible in real scenarios.
In this paper, we find the performance drops when detecting unseen videos may be highly related to the different spatiotemporal patterns existed in different videos, and based on this observation, we further propose the Latent Spatiotemporal Adaptation~(LAST) framework by recovering the spatiotemporal clues of unseen target videos in latent space to achieve generalized face forgery video detection.

\subsection{Unsupervised Domain Adaptation}
Unsupervised domain adaptation~(UDA) focuses on how to transfer knowledge from a labeled source domain to an unlabeled target domain. Many UDA methods formulate this as a distribution alignment problem. Some researchers also introduce designed auxiliary tasks to address this issue, such as adversarial learning~\cite{DBLP:conf/cvpr/TzengHSD17}
, and image reconstruction or style transfer~\cite{DBLP:conf/cvpr/MurezKKRK18,DBLP:conf/eccv/GhifaryKZBL16}. By designing the above auxiliary tasks, the generalization of the main task could be improved, also consistent with the goal of multi-task learning~\cite{argyriou2006multi}.
%For face forgery detection task, we can formulate the data generated from different methods or source videos as from different domains, since they exist distribution gaps.
There are some existing detectors formulate the task as domain adaptation problem~\cite{yin2024improving} to improve the generalization. But current methods either require data from different domains with labels, or can not adapt to unseen domains.
%With the generation methods still evolving, a desired forgery detector should generalize well when detecting unseen videos.
In this paper, we regard the known forgery data as labeled source domain and the unseen videos to detect as unlabeled target domain, and formulate the task as a UDA problem. %Since the target videos for testing are easily accessible in real scenarios, our formulation may be more appropriate and practical to achieve a generalized detector. 
Based on our observation above, we first model the spatiotemporal representation by incorporating CNN for local spatial feature cascaded with vision transformer for long-term memory, then aim to recover the spatiotemporal clues of unseen target videos in latent space to bridge the domain gaps.

\subsection{Self-Supervised Learning}
Self-supervised learning~(SSL) aims to automatically generate supervision from the data without labels~\cite{DBLP:journals/corr/abs-2111-06377,pathak2016context}.
The most common strategy for SSL is designing auxiliary pretext tasks to introduce supervised signals, such as inpainting~\cite{pathak2016context} and reconstruction learning~\cite{DBLP:journals/corr/abs-2111-06377}. 
Contrastive learning has achieved impressive performance on self-supervised representation learning~\cite{tian2020contrastive,wang2022cross}, which aims to pull close the positive pairs and push away the negative pairs in latent space.
Recent studies have also demonstrated the impressive self-supervised representation capacity of the vision transformer~(ViT)~\cite{DBLP:conf/iclr/DosovitskiyB0WZ21,DBLP:journals/corr/abs-2203-01265}.
There exist some self-supervised forgery detectors, but they either focus on specific forgery patterns~\cite{DBLP:journals/corr/abs-2203-12208,DBLP:conf/cvpr/LiL19c,zhang2022unsupervised} or the cross-modal mismatch~\cite{DBLP:journals/corr/abs-2203-01265,DBLP:conf/cvpr/HaliassosMPP22}, which still focus on specific forgery data and may lead to limited generalization facing unseen videos. 
Instead, to eliminate the influence introduced by specific forgery videos, we propose to pre-train our CNN and transformer on real-only face videos with two simple yet effective self-supervised tasks to learn the common spatiotemporal representation of face videos as initialization and keep them frozen during adaptation. %The self-supervised tasks include reconstruction and contrastive learning in latent space: reconstruction learning could force the learned representations to contain more spatiotemporal clues of original videos, and contrastive learning could explore the intra- and inter-video connections, both should benefit common spatiotemporal learning.

\begin{figure*}[t]
   \centering
   \includegraphics[width=\linewidth]{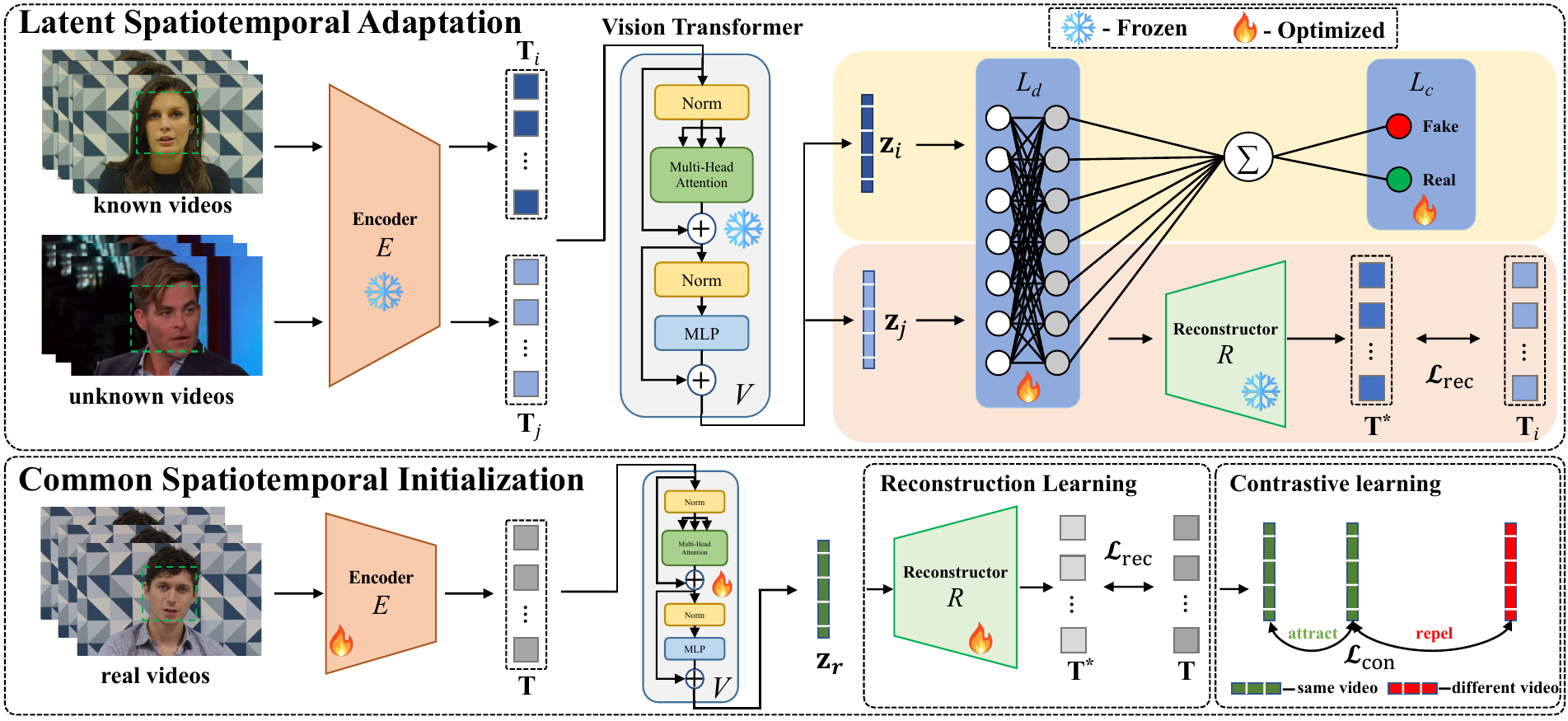}
   \caption{\label{figure:pipeline} The whole pipeline of our proposed method, including the latent spatiotemporal adaptation stage~(top) and the common spatiotemporal initialization stage~(bottom).}
   \vspace{-.5em}
   %The input videos are first extracted into local spatial features by CNN encoder and then cascade by a vision transformer to learn the spatiotemporal representation. During adaptation~(top), the representations are fed into the linear head to perform the usual classification task on labeled source videos and also recover the spatiotemporal clues of unlabeled target videos. To eliminate the influence of specific forgery videos when optimizing the latent space, we further propose the spatiotemporal initialization~(bottom) on real-only videos to learn the common spatiotemporal representation and keep them frozen during adaptation. }
\end{figure*}

\section{Proposed Method}\label{sec:method}

%In this section, we discuss our proposed method~(LAST) from three aspects: the spatiotemporal representation learning, latent spatiotemporal adaptation, and common spatiotemporal initialization, as illustrated in Fig.~\ref{figure:pipeline}. 
%The following subsections present the details of each part.

\subsection{Spatiotemporal Representation Learning}
We aim to leverage the spatiotemporal clues in videos to achieve generalized face forgery detection. Thus, the first step is to model the spatiotemporal representation of video effectively. We first choose to model the spatiotemporal information in latent space instead of pixel space instead of the redundant pixel space~\cite{DBLP:journals/corr/abs-2111-06377}.
Inspired by the advantages of CNN in local reception field~\cite{DBLP:conf/cvpr/HeZRS16} and vision transformer in long-term memory~\cite{DBLP:conf/iclr/DosovitskiyB0WZ21,ramachandran2019stand}, we develop a framework by first using a simple CNN to extract the local spatial feature of each frame and then cascading a vision transformer to learn the long-term spatiotemporal representation, illustrated in Fig.~\ref{figure:pipeline}. 
%In this way, we assume the learned representation should contain both local single frame and global long-term spatiotemporal clues from video, which should be useful for the whole task. Each part will be described in detail below.

\subsubsection{Local Receptive Spatial Feature}
We first extract the local spatial feature of each frame from the video by a simple CNN encoder ${E}(\cdot)$. Due to the inherent advantage of CNN on local reception field~\cite{DBLP:conf/cvpr/HeZRS16}, the extracted feature should contain more useful local spatial information than the redundant pixel space~\cite{DBLP:journals/corr/abs-2111-06377}, which could benefit our task.

For the sampled clip~$\mathbf{X} = \left\{ \mathbf{x}_i\right\}_{i=1}^{n}$, we introduce a simple CNN encoder ${E}(\cdot)$ to extract the spatial feature of each single frame~$\mathbf{x}_i$, which projects the frame sequence into the spatial feature sequence $\mathbf{T}=\left\{ \mathbf{t}_i\right\}_{i=1}^{n}$, formulated as follows:
\begin{equation}
    \mathbf{T} = {E}(\mathbf{X}) = \operatorname{Conv}(\mathbf{X},\Theta_e) = \operatorname{Conv}(\left\{ \mathbf{x}_i\right\}_{i=1}^{n}, \Theta_e) = \left\{ \mathbf{t}_i\right\}_{i=1}^{n},
\end{equation}
where the~$\{\Theta_e\}$ is the parameters of the encoder ${E}(\cdot)$ and the dimension of extracted local spatial feature~$\mathbf{t}_i$ is 256.

\subsubsection{Long-term Spatiotemporal Representation}
%Based on the extracted local spatial feature sequence~$\mathbf{T}$, we cascade a vision transformer~${V}(\cdot)$ to learn their long-term connections thus obtaining the spatiotemporal representation. 
Some previous findings~\cite{DBLP:conf/iclr/DosovitskiyB0WZ21,ramachandran2019stand} demonstrate vision transformer's superiority in long-term memory due to the attention mechanism. Thus, we cascade a vision transformer to learn the long-term spatiotemporal representation.
Specifically, we first reshape local spatial feature $\mathbf{t}_i$ into $16 \times 16$ tokens and employ a trainable linear projection matrix~$\mathbf{W}$ with positional embedding~$\mathbf{E}_{pos}$ to get the input feature sequence and feed them into our vision transformer~${V}(\cdot)$ to learn the spatiotemporal representation~$\mathbf{z}$ of each video clip. Our vision transformer consists of $K$ standard Transformer Encoder blocks, which contains a multi-head self-attention~(MSA) block and an MLP header with the commonly-used LayerNorm~(LN) before them. We use the GELU function as the activation function and the learning process can be formulated as follows:
\begin{align}
    & \mathbf{z}_0 = \mathbf{W} \cdot \mathbf{F} + \mathbf{E}_{pos} = \mathbf{W} \cdot \left[ \mathbf{t}_1, \cdots, \mathbf{t}_n\right]^{\mathrm{T}} + \mathbf{E}_{pos},\\
    & \mathbf{z}_k = \operatorname{MSA}(\operatorname{LN}(\mathbf{z}_{k-1})) + \mathbf{z}_{k-1}, \qquad k=1\ldots K
\end{align}
Further, we denote the trainable parameters of our vision transformer as~$\{\Theta_v\}$ and naturally use the output feature of last layer~$\mathbf{z}_K$ with dimension of 768 as the final learned spatiotemporal representation. In the following sections, we use $\mathbf{z}$ to denote the $\mathbf{z}_K$ for simplicity unless stated otherwise.

\subsection{Latent Spatiotemporal Adaptation}
Equipped with the designed representation learning method above, we have a desired spatiotemporal latent space that contains more useful clues than pixel space. Then the question is how to achieve a generalized detector that can effectively adapt to unknown videos. 
Based on our findings in Fig.~\ref{figure:mot}, the different spatiotemporal patterns in different videos may be the key to generalize.
%Thus, we further propose the latent spatiotemporal adaptation paradigm to improve the model's generalization to unknown videos. 
We further propose to perform the usual forgery detection task on known videos and also recover the spatiotemporal clues on unknown videos in latent space in a semi-supervised manner. By reconstructing the spatiotemporal clues as auxiliary task, the distribution gap of known and unknown videos could be mitigated, thus improving the generalization. This paradigm is also consistent with the idea of multi-task learning~\cite{argyriou2006multi}. And the unlabeled target videos are usually easily accessible in real scenarios, which also makes our method more practical.

%Since our adaptation paradigm includes the known and unknown video distributions, and usually the known videos' labels are accessible while the unknown videos' labels are missing, 
Specifically, we regard the known videos as labeled source domain and the unknown videos as unlabeled target domain to formulate this problem as a \textit{unsupervised domain adaptation} task.
First, we present a brief formulation of {unsupervised domain adaptation} problem: we define a domain as a probability distribution $\mathcal{D}$ on $\mathcal{X} \times \mathcal{Y}$, where $\mathcal{X}$ is the data space and $\mathcal{Y}$ is the label space. Then denote the source domain by $\mathcal{D}_s$ and target domain by $\mathcal{D}_t$, where $\mathcal{D}_s \neq \mathcal{D}_t$.
Given the labeled sample from one specific source domain $\mathbf{D}_s = \left\{ (x_i^s, y_i^s) \right\}_{i=1}^{n_s} \sim \mathcal{D}_s$ and unlabeled sample from one target domain $\mathbf{D}_t = \left\{
x_i^t \right\}_{i=1}^{n_t} \sim \mathcal{D}_t^{\mathcal{X}}$,
our goal is to find a function: $\mathbb{f}_t: \mathcal{X} \rightarrow \mathcal{Y}$ on $\mathbf{D}_t$ which minimizes the discrepancy between predicted label distribution $\mathbb{f}_t \left( \mathbf{D}_t \right)$ and ground truth label distribution $\mathcal{D}_t^\mathcal{Y}$. 
Compared to $\mathbb{f}_t$, it is easier to find a function: $\mathbb{f}_s: \mathcal{X} \rightarrow \mathcal{Y}$ on $\mathbf{D}_s$ by supervised learning since their labels are accessible. 
Based on this and from the perspective of representation learning, the previous goal is equivalent to finding a function $\mathbb{g}: \mathcal{X} \rightarrow \mathcal{Z}$, where $\mathcal{Z}$ is the latent space, such that the discrepancy between $\{x^{s}_{i}\} \in \mathbf{D}_s$ and $\{x^{t}_{i}\} \in \mathbf{D}_t$ is minimized in $\mathcal{Z}$. 
Then we could easily optimize another function $\mathbb{f}: \mathcal{Z} \rightarrow \mathcal{Y}$ based on the labeled sample from $\mathcal{D}_s$ in a supervised way to predict the label from latent space, thus achieve generalization on $\mathbf{D}_t$.

Based on the analysis above, our goal is to find the function $\mathbb{g}(\cdot)$ from data space to latent space to bridge the distribution gap in latent space, and also the function $\mathbb{f}(\cdot)$ from latent space to label space to predict accurately. 
Thus, latent space learned by~$\mathbb{g}(\cdot)$ should satisfy the following two criteria: \emph{(1) preserve the discriminability on the labeled source domain} and \emph{(2) contain enough useful information that can be adapted to unlabeled target domain}. The criteria (1) can ensure the detector preserves its discriminability and the criteria (2) can ensure the shared latent space contains clues from unlabeled target domain to generalize. It is easy to achieve criteria (1) by supervised learning on labeled source domains and the remaining challenge is how to ensure the latent space contains useful information of unlabeled target domain to generalize. 
%Based on our findings above, we assume the different spatiotemporal patterns in different forgery videos may be the reason leading to distribution gap. 
To tackle this issue, we propose to recover the spatiotemporal clues of unlabeled target domain to bridge the distribution gap in latent space, which could force the learned latent space contain more spatiotemporal clues of unlabeled target domain. % leading to better generalization on unknown forgery videos.

As shown on the top of Fig.~\ref{figure:pipeline}, we propose the paradigm of latent spatiotemporal adaptation. Specifically, we add two linear layers after the vision transformer: the adaptive layer~${L}_{d}(\cdot)$ aims to learn a shared and adaptive latent space, and the classification layer~${L}_{c}(\cdot)$ is the usual supervised head for forgery classification task. The adaptive latent space learned by~${L}_{d}(\cdot)$ is optimized by recovering the local spatial features from spatiotemporal representation in latent space, which could force the shared latent space contain more useful spatiotemporal clues of target videos leading to improved generalization.
We define $f_\textsubscript{cls} : \mathcal{X} \rightarrow \mathcal{Y}$ as the supervised classification pipeline on the labeled source domain, and $f_\textsubscript{rec} : \mathcal{X} \rightarrow \mathcal{X}$ as the unsupervised reconstruction pipeline on the unlabeled target domain. %where the $\mathcal{X}$ is the data space and $\mathcal{Y}$ is the label space. 
By optimizing the two pipelines jointly, the shared latent space learned by ${L}_{d}(\cdot)$ from different domains are bridged together, serving as the function $\mathbb{g}(\cdot)$, and the classifier ${L}_{c}(\cdot)$ can preserve its discriminability on classification, serving as the function $\mathbb{f}(\cdot)$.

Let $ \Theta = \left \{ \Theta_{d}, \Theta_{c} \right \}$ denote the parameters of ${L}_{d}(\cdot)$ and ${L}_{c}(\cdot)$ respectively.
For $f_\textsubscript{cls}$, we optimize it in the usual supervised manner on labeled source videos from latent space to label space.
For $f_\textsubscript{rec}$, we reconstruct the local spatial feature sequence~$\mathbf{T}^*$ from learned spatiotemporal representation~$\mathbf{z}$ on unlabeled target videos in latent space by an introduced CNN reconstructor~$R(\cdot)$. %Here, the shared latent space learned by~${L}_{d}(\cdot)$ serves as the desired and generalized latent space $\mathcal{Z}$.
%Thus, the objective of our SAR can be described below. 
Given labeled source domain $\mathbf{D}_{s} = \left\{  \left( \mathbf{x}_i^s, \mathbf{y}_i^s \right) \right\}_{i=1}^{n_s}$ and unlabeled target domain $\mathbf{D}_t = \left\{ \left( \mathbf{x}^t_i \right) \right\}_{i=1}^{n_t}$, we can formulate the two pipelines as follows:
\begin{gather}
    \mathcal{L}^{n_s}_\textsubscript{cls} (  \{ \Theta_{d},\Theta_{c} \} ) = \sum_{i=1}^{n_s} \mathcal{L}_\textsubscript{cls} \left( f_\textsubscript{cls} (\mathbf{z}_i^s; \{ \Theta_{d},\Theta_{c} \}), \mathbf{y}_i^s\right), \\
    \mathcal{L}^{n_t}_\textsubscript{rec} (  \{ \Theta_{d} \} ) = \sum_{i=1}^{n_t} \mathcal{L}_\textsubscript{rec} \left( f_\textsubscript{rec} (\mathbf{z}_i^t; \{ \Theta_{d}, \Theta_{r}\}), \mathbf{T}_i^t\right),
\end{gather}
where $\{ \mathbf{z}_i^s, \mathbf{z}_i^t \}$ is the learned spatiotemporal representations of $\{ \mathbf{x}_i^s, \mathbf{x}_i^t \}$ respectively, $\{\mathbf{T}_i^{t}\}$ is the local spatial feature sequence of $\{\mathbf{x}_i^t\}$, $\{ \Theta_{r} \}$ is the parameters of the reconstructor~$R(\cdot)$, $\mathcal{L}_\textsubscript{cls}$ is the binary cross-entropy loss and we choose $\ell_{1}$ norm as the distance function for $\mathcal{L}_\textsubscript{rec}$.
Noticing that the parameters $\left \{  \Theta_{d}, \Theta_{c} \right \}$ are jointly optimized in $f_\textsubscript{cls}$, while only the $\{ \Theta_{d} \}$ is optimized in $f_\textsubscript{rec}$.% and the~$\{ \Theta_{r} \}$ is frozen after initialization, described in detail in following section.

Finally, our goal is to minimize the following objective:
\begin{equation}
  \label{eq:last}
   \mathcal{L}_\textsubscript{last} = \lambda \mathcal{L}^{n_s}_\textsubscript{cls} + (1-\lambda) \mathcal{L}^{n_t}_\textsubscript{rec},
\end{equation}
where $0 \leq \lambda \leq 1$ is a hyper-parameter that controls the trade-off between the supervised classification and unsupervised reconstruction pipelines. 

\subsection{Common Spatiotemporal Initialization}
Based on our previous findings, we observe different forgery videos exist different spatiotemporal patterns which may cause performance drop. Thus, if we learn the spatiotemporal latent space from specific forgery videos, the learned latent space may also contain specific forgery clues that may undermine the adaptation process. To eliminate the influence of specific forgery videos, we propose to initialize the spatiotemporal latent space learned by our CNN including~$\{E(\cdot), R(\cdot)\}$ and vision transformer~${V(\cdot)}$ by pre-training on real-only videos to learn the common spatiotemporal patterns and keep them frozen during adaptation. Since there is no forgery video included, the initial latent space which contains the common spatiotemporal patterns of real face videos should generalize better. To this end, we employ two simple yet effective self-supervised tasks in latent space, as shown at the bottom of Fig.~\ref{figure:pipeline} and described as follows.

\subsubsection{Reconstruction Learning}
%The reconstruction learning is a widely used auxiliary task in self-supervised learning and some previous work also demonstrates it could benefit video spatiotemporal representation learning~\cite{li2022locality,weng2021event}.
We assume that by reconstructing the spatiotemporal clues, the latent space should contain more information for both spatial structure and temporal connection~\cite{li2022locality,weng2021event}. 
To this end and incorporate with the adaptation process, we introduce the reconstruction task by recovering the local spatial feature from learned spatiotemporal representation in latent space, which could force the learned spatiotemporal representation contain more spatiotemporal clues of input real videos. Specifically, we introduce one convolutional reconstructor $R(\cdot)$ with parameters $\{\Theta_{r}\}$ which takes the learned spatiotemporal representation $z$ as input and output $\mathbf{T}^*$ to recover the original spatial feature sequence $\mathbf{T} = \left\{ \mathbf{t}_i \right\}_{i=1}^{n}$, formulated as follows:
\begin{equation}
    \label{eq:rec}
    \mathcal{L}_\textsubscript{rec} = \left\| \mathbf{T}^{*} - \mathbf{T} \right\| = \frac{1}{n} \sum_{i=1}^{n}\left\| \mathbf{t}_{i}^{*} - \mathbf{t}_{i} \right\|,
\end{equation}
where $n$ is the frame and token sequence length, and we choose $\ell_{1}$ norm as our distance function.

\subsubsection{Contrastive Learning}
%Some previous work has demonstrated that mining both in intra-video and inter-video connections could benefit the spatiotemporal learning on videos~\cite{qian2021spatiotemporal}.
We aim to  explore both intra- and inter-video relationship to learn the common spatiotemporal representation of real videos by contrastive learning~\cite{qian2021spatiotemporal}.
%Thus, we employ contrastive learning to explore both intra- and inter-video relationship to learn the common spatiotemporal representation of real videos.
To this end, we first define the similarity of two different spatiotemporal representations~$\left( \mathbf{z}_i, \mathbf{z}_j \right)$, formulated as follows:
\begin{equation}
    \label{eq:sim}
    \operatorname{sim} \left(\mathbf{z}_i, \mathbf{z}_j\right) = \frac{\mathbf{z}_i \cdot \mathbf{z}_j}{\mathop{\max}(\left\| \mathbf{z}_i \right\|_{2} \cdot \left\| \mathbf{z}_j \right\|_{2}, \epsilon)},
\end{equation}
where $\epsilon$ is set to $1e-8$. For one input clip $\mathbf{X} = \left\{ \mathbf{x}_i\right\}_{i=1}^{n}$, we consider the representation learned from the clips sampled from the same video as positive pairs~$\mathbf{z}^+$, while those sampled from different videos as negative pairs~$\mathbf{z}^-$, which could both explore the intra- and inter-video spatiotemporal relations. Thus, we can formulate the contrastive loss~$\mathcal{L}_\textsubscript{con}$ within mini-batch samples as follows:
%\begin{small}
\begin{equation}
    \label{eq:con}
    \mathcal{L}_\textsubscript{con}=-\log \frac{\exp \left( \operatorname{sim}\left(\mathbf{z}, \mathbf{z}^+ \right) / \tau\right)}{\exp \left(\operatorname{sim} \left(\mathbf{z}, \mathbf{z}^+ \right) / \tau\right)+\sum \exp \left( \operatorname{sim} \left(\mathbf{z}, {\mathbf{z}^-} \right) / \tau\right)},
\end{equation}
%\end{small}
where the temperature $\tau$ is set to 0.5. Specifically, for each input video, we sample other clips with the same fixed length~$n$ from a different random offset as positive pairs. Moreover, in each pretraining mini-batch of~$\mathrm{M}$ videos, we consider the clips from other~$(\mathrm{M} - 1)$ videos as negative pairs, which means there are total $\left( \mathrm{M} - 1 \right) \times \mathrm{M}$ negative pairs.

Therefore, the total loss function for our self-supervised pretraining can be formulated as follows:
\begin{gather} 
    \label{eq:init}
    \mathcal{L}_\textsubscript{init} = \lambda_1 \mathcal{L}_\textsubscript{con} + \lambda_2 \mathcal{L}_\textsubscript{rec},
\end{gather}
where $\lambda_1$ and $\lambda_2$ are the hyper-parameter weights to balance the contrastive and reconstruction loss. Thus,  the learned~$\{ \Theta_{e}, \Theta_{r}, \Theta_{v}\}$ are not influenced by any specific forgery clues and we keep them frozen during adaption to further enhance the generalization.

\begin{table*}
\centering
%\tiny
\small
\caption{Cross-dataset generalization comparisons with the state-of-the-art detectors when trained on FF++. 
The methods above the line are evaluated at image-level while others are at video-level.}\label{tab:cross-dataset}
  %\resizebox{\linewidth}{!}{
  \setlength{\tabcolsep}{4mm}{
  \begin{tabular}{l|cc|cc|cc}
    %\toprule
    \hline
    \multirow{2}{*}{Method} &
        \multicolumn{2}{c|}{DFDC} &
        \multicolumn{2}{c|}{Celeb-DF} &
        \multicolumn{2}{c}{DFD}\\
        \hhline{~|--||--||--|}
        & {ACC(\%)} & {AUC(\%)} & {ACC(\%)} & {AUC(\%)} & {ACC(\%)} & {AUC(\%)}\\
    \hline
    Xception~\cite{Rssler2019FaceForensics} & 57.76 & 60.70 & 71.87 & 74.67 & 73.86 & 80.09\\
    LTW~\cite{DBLP:conf/aaai/Sun0YGLSJ21} & 63.10 & 69.00 & 63.40 & 64.10 & - & -\\
    F$^{3}$-Net~\cite{DBLP:conf/eccv/QianYSCS20} & - & 72.88 & - & 71.21 & - & 86.10\\
    %RECCE~\cite{cao2022end} & - & 69.06 & - & 68.71 & - & -\\
    %HFI-Net~\cite{DBLP:journals/tifs/MiaoTCYG22} & - & 73.65 & - & 83.29 & - & -\\
    MultiAtt~\cite{zhao2021multi} & 60.68 & 62.91 & 65.34 & 68.19 & 71.61 & 74.09\\
    RFM~\cite{wang2021representative} & - & - & - & 74.17 & - & 89.30\\
    MPSM~\cite{DBLP:conf/aaai/ChenYCDLJ21} & 72.29 & 76.61 & 76.86 & 78.37 & 84.53 & 89.29 \\
    PEL~\cite{gu2022exploiting} & 62.32 & 65.87 & 66.54 & 70.29 & 73.31 & 77.61 \\
    Capsule~\cite{DBLP:journals/corr/abs-1910-12467} & 59.55 & 64.49 & 70.08 & 71.79 & 71.67 & 76.75\\
    Face X-Ray~\cite{li2020face} & 62.37 & 67.32 & 62.69 & 68.22 & 76.19 & 81.24\\
    SPSL~\cite{liu2021spatial} & 62.26 & 67.97 & 72.44 & 75.17 & 74.34 & 82.64\\
    SLADD~\cite{DBLP:journals/corr/abs-2203-12208} & - & 68.80 & - & 74.75 & - & 87.66\\
    CORE~\cite{ni2022core} & - & 62.51 & - & 72.42 & - & 91.35\\
    GFFD~\cite{DBLP:conf/cvpr/LuoZY021} & 55.66 & 60.38 & 62.51 & 69.57 & 76.63 & 82.59\\
    CADDM~\cite{dong2023implicit} & 53.99 & 62.88 & 60.58 & 71.91 & 74.66 & 85.65\\
    MCS-GAN~\cite{DBLP:journals/tmm/XiaoLYLMG24} & - & 48.90 & - & 54.30 & - & -\\
    ID$_3$~\cite{yin2024improving} & 64.68 & 70.85 & 73.21 & 76.81 & 80.38 & 87.59\\
    NiCL~\cite{qiao2024deepfake} & - & 62.30 & - & 64.69 & - & 90.78 \\
    \arrayrulecolor{black}\hline
    Two-branch \cite{DBLP:conf/eccv/MasiKMGA20} & 67.37 & 70.18 & 72.63 & 75.79 & 83.42 & 86.19\\
    %CNN-GRU~\cite{DBLP:conf/cvpr/SabirCJAMN19} & - & 68.90 & - & 69.80 & - & -\\
    %Multi-task~\cite{nguyen2019multi} & - & 68.10 & - & 75.70 & - & -\\
    %CNN-aug~\cite{DBLP:conf/cvpr/WangW0OE20} & - & 72.10 & - & 75.60 & - & -\\
    S-MIL~\cite{DBLP:conf/mm/LiLCMHWXL20} & 64.74 & 66.79 & 73.29 & 75.86 & 84.86 & 86.39\\
    Patch-based~\cite{DBLP:conf/eccv/ChaiBLI20} & - & 65.60 & - & 69.60 & - & -\\
    DSP-FWA~\cite{DBLP:conf/cvpr/LiYSQL20} & - & 67.30 & - & 69.50 & - & -\\
     STIL~\cite{DBLP:conf/mm/GuCYDLHM21} & 65.95 & 67.22 & 74.67 & 76.17 & 85.84 & 87.21\\
    %HCIL~\cite{DBLP:conf/eccv/GuYCDM22} & - & 69.21 & - & 79.00 & - & -\\
    %LipForensics~\cite{DBLP:conf/cvpr/HaliassosVPP21} & & 73.50 & & 82.40 & - & -\\
    TD3DCNN~\cite{DBLP:conf/ijcai/ZhangLL0G21} & 82.21 & 55.02 & 66.02 & 57.32 & - & -\\
    DIANet~\cite{DBLP:conf/ijcai/HuXWLW021} & 65.95 & 68.65 & 70.83 & 73.94 & 77.56 & 79.91\\
    STDT~\cite{DBLP:journals/corr/abs-2207-06612} & 83.87 & 66.99 & 68.92 & 69.78 & - & -\\
    DIL~\cite{DBLP:conf/aaai/GuCYDLM22} & 70.93 & 72.52 & 81.18 & 82.82 & 86.22 & 88.15\\
    ISTVT~\cite{zhao2023istvt} & - & 74.20 & - & 84.10 & - & - \\
    NoiseDF~\cite{wang2023noise}  & 59.87 & 63.89 & 70.10 & 75.89 & - & - \\
    LAST~(ours) & \textbf{85.28} & \textbf{74.88} & \textbf{81.66} & \textbf{86.43} & \textbf{86.81} & \textbf{92.01} \\
    %\bottomrule
    \arrayrulecolor{black}\hline
  \end{tabular}}
  \vspace{-.25em}
\end{table*}

\section{Experiments}
\label{sec:exp}

\subsection{Experimental Settings}
\myPara{Datasets.}
For adaptation and evaluation, we take four widely used face forgery video datasets: FaceForensics++~(FF++)~\cite{Rssler2019FaceForensics}, Celeb-DF(v2)~\cite{DBLP:conf/cvpr/LiYSQL20}, DeepFake Detection~(DFD)~\cite{googledfd19} and DeepFake Detection Challenge~(DFDC)~\cite{dolhansky2019the}. FF++ is generated by four different methods: Deepfakes (DF)~\cite{deepfakes}, Face2Face~\cite{thies2016face2face:}~(F2F), FaceSwap (FS)~\cite{Faceswap} and NeuralTextures~\cite{thies2019deferred}~(NT). It provides three different compression levels and we use its low compressed dataset~(c23). Celeb-DF(v2) includes high-quality fake videos covering different ages, genders, and ethics, which are generated by the improved DeepFake algorithm~\cite{DBLP:conf/cvpr/LiYSQL20}. DFD is recorded from hiring actors and generated by public available methods. DFDC has a variety of manipulation methods, it is a challenging dataset for existing detectors. We randomly sample 1/10 unlabeled target videos compared to labeled source videos for adaptation, and regard the training forgery dataset as labeled source domain while the testing dataset as unlabeled target domain.

For the initialization, we use two real face video datasets: VoxCeleb2~\cite{DBLP:conf/interspeech/ChungNZ18} and AVSpeech~\cite{DBLP:journals/tog/EphratMLDWHFR18} to learn the common spatiotemporal representation of face videos. 

\myPara{Evaluation metrics.}~Following recent work, we report the most commonly used metrics~\cite{Rssler2019FaceForensics,li2020face,DBLP:conf/wifs/AfcharNYE18} to compare with prior works, including the Accuracy~(ACC(\%)), the Area Under the Receiver Operating Characteristic Curve~(AUC(\%)), and the Equal Error Rate~(EER(\%)). All metrics are evaluated at video-level unless stated otherwise.
%For image-level detectors, we calculate the mean output of multiple extracted frames. All the metrics are evaluated and reported at video-level.

\myPara{Implementation details.}~We extract frames with OpenCV, employ face alignment with~\cite{facealign}, reshape facial frames into 224$\times$224, and sample clips with continuous 20 frames for each video from a random offset as input facial frame sequence. The CNN encoder $E(\cdot)$ consists of three $3\times3$ convolutional layers with 64, 128 and 256 channels following by ReLU activations and one average pooling layer, the CNN reconstructor $R(\cdot)$ consists of one convolutional layer, and the vision transformer uses a 12-block ViT-Base-16 \cite{DBLP:conf/iclr/DosovitskiyB0WZ21} that outputs a 768 dimension spatiotemporal representation for each video. 
The linear layer ${L}_{d}(\cdot)$ has the same output dimension with the output of vision transformer, and the ${L}_{c}(\cdot)$ output 2 dimension vectors representing real and fake class.
We set batch size as 64, use Adam optimizer with an initial learning rate of $5e-4$ and a weight decay of $1e-4$, employ a warm-up scheduler to adjust the learning rate which is empirically set to $1e-5$. 
We empirically set $\lambda=0.5$ in Eq.~(\ref{eq:last}), and $\lambda_1 = 1.0$ and $\lambda_2 = 0.5$ in Eq.~(\ref{eq:init}).
Furthermore, we pretrain models for 100 epochs and then employ the adaptation for 10 epochs, which could further support the flexibility and efficiency of our method.

\begin{table}[ht]
\small
   \centering
   \caption{{Architectures and trainable parameters for the usual face forgery video detection task. }
   }\label{tab:params-arch}
   \resizebox{.9\linewidth}{!}{
       \begin{tabular}{llrccc}
          %\toprule
          \hline
              {Method} & Arch & {\#params} \\
              %{Clean} & {CS} & {CC} & {BW} & {GNC} & {GB} & {PX} & {VC} & \textit{Avg}\\
              %Train & Test&&&&&&\\
              %\midrule
              \hline
              {MultiAtt} \cite{zhao2021multi} & CNN+Attention & 41.7M \\
              {Two-branch} \cite{DBLP:conf/eccv/MasiKMGA20} & CNN+LSTM & 67.1M \\
              {MPSM} \cite{DBLP:conf/aaai/ChenYCDLJ21} & CNN+Attention & 46.3M \\
              {PEL} \cite{gu2022exploiting} & CNN & 33.7M \\
              {STIL} \cite{DBLP:conf/mm/GuCYDLHM21} & CNN & 32.5M \\
              {S-MIL} \cite{DBLP:conf/mm/LiLCMHWXL20} & CNN & 42.2M \\
              {DIL} \cite{DBLP:conf/aaai/GuCYDLM22} & CNN & 41.4M \\
              {DIANet} \cite{DBLP:conf/ijcai/HuXWLW021} & CNN+Attention & 32.7M \\
              SeqFakeFormer \cite{shao2022detecting} & CNN+Transformer& 104M \\
              {LipForensics} \cite{DBLP:conf/cvpr/HaliassosVPP21} & CNN+MS-TCN & 24.8M\\ %& 97.4\\
              {FTCN} \cite{DBLP:conf/iccv/ZhengB0ZW21} & 3D-CNN+Transformer & 26.6M \\ %& 98.8\\
              {RealForensics} \cite{DBLP:conf/cvpr/HaliassosMPP22} & CNN+CSN & 21.4M \\% & 99.5\\
              AltFreezing \cite{wang2023altfreezing} & 3D-CNN & 27.2M\\
              \hline
              {LAST~(ours)} & CNN+Transformer & \textbf{4.5M} \\%& 99.2\\
          %\bottomrule
          \hline
          \vspace{-2em}
       \end{tabular}
   }
\end{table}

\begin{table}[ht]
\centering
\caption{Cross-manipulation generalization results~(video-level AUC(\%)) on four subsets of FF++ dataset where trained on one subset and tested on others.}\label{tab:cross-manipulation}
  \resizebox{\linewidth}{!}{
  \begin{tabular}{l|c|cccc|c}
    %\toprule
    \hhline{|-------|}
    \multirow{2}{*}{Approach} &
     \multirow{2}{*}{Train Dataset} &
      \multicolumn{4}{c|}{Test Dataset} \\
      %\cmidrule(lr){3-6}
      \hhline{~~|----|~}
        & &
        \multicolumn{1}{c}{DF} &
        \multicolumn{1}{c}{F2F} &
        \multicolumn{1}{c}{FS} &
        \multicolumn{1}{c|}{NT} &
        \multicolumn{1}{c}{\textit{Avg}} \\
    %\midrule
    \hhline{|-------|}
    Xception~\cite{Rssler2019FaceForensics} & \multirow{7}{*}{DF} & \multirow{7}{*}{-} & 58.81 & 64.79 & 59.69 & 61.10  \\
    MesoNet~\cite{DBLP:conf/wifs/AfcharNYE18} & & & 52.27 & 62.17 & 53.53 & 55.99\\
    LRNet~\cite{sun2021improving} & & & 65.84 & 65.66 & 64.09 & 65.20\\
    CNN-LSTM~\cite{guera2018deepfake} & & & 62.43 & 58.63 & 64.18 & 61.75\\
    Freq-SCL~\cite{DBLP:conf/cvpr/LiXLW021} & & & 58.90 & 66.87 & 63.61 & 63.13\\
    MultiAtt~\cite{zhao2021multi} & & & 66.41 & 67.33 & 66.01 & 66.58\\
    LAST~(ours) & & & {\textbf{71.72}} & \textbf{70.72} & \textbf{66.79} & \textbf{69.74}\\%99.75
    %\midrule
    \hhline{|-------|}
    Xception~\cite{Rssler2019FaceForensics} & \multirow{7}{*}{F2F} & 66.39 & \multirow{7}{*}{-} & 56.58 & 57.59 & 60.19 \\
    MesoNet~\cite{DBLP:conf/wifs/AfcharNYE18} & & 54.54 & & 52.26 & 54.53 & 53.78\\
    LRNet~\cite{sun2021improving} & & 81.93 & & 62.66 & \textbf{79.53} & \textbf{74.71}\\
    CNN-LSTM~\cite{guera2018deepfake} & & 61.11 & & 53.37 & 60.56 & 58.35\\
    Freq-SCL~\cite{DBLP:conf/cvpr/LiXLW021} & & 67.55 &  & 55.35 & 66.66 & 63.19\\
    MultiAtt~\cite{zhao2021multi} & & 73.04 & & 65.10 & 71.88 & 70.01\\
    LAST~(ours) & & {\textbf{82.25}} & & \textbf{67.78} & {69.95} & {73.33} \\%99.57
    %\midrule
    \hhline{|-------|}
    Xception~\cite{Rssler2019FaceForensics} & \multirow{7}{*}{FS} & 80.00 & 56.65 & \multirow{7}{*}{-} & 53.42 & 63.36 \\
    MesoNet~\cite{DBLP:conf/wifs/AfcharNYE18} & & 63.12 & 48.74 & & 47.74 & 53.20\\
    LRNet~\cite{sun2021improving} & & 73.84 & \textbf{68.54} & & 49.26 & 63.88\\
    CNN-LSTM~\cite{guera2018deepfake} & & 54.18 & 46.67 & & 43.28 & 48.04\\
    Freq-SCL~\cite{DBLP:conf/cvpr/LiXLW021} & & 75.90 & 54.64 & & 49.72 & 60.09\\
    MultiAtt~\cite{zhao2021multi} & & \textbf{82.33} & 61.65 & & 54.79 & 66.26\\
    LAST~(ours) & & {80.37} & {64.01} & & {\textbf{75.30}} & \textbf{73.23} \\ %99.81
    %\midrule
    \hhline{|-------|}
    Xception~\cite{Rssler2019FaceForensics} & \multirow{7}{*}{NT} & 69.94 & 67.88 & 57.59 & \multirow{7}{*}{-} & 65.14 \\
    MesoNet~\cite{DBLP:conf/wifs/AfcharNYE18} & & 55.94 & 55.98 & 43.00 & & 51.64\\
    LRNet~\cite{sun2021improving} & & 80.46 & \textbf{89.39} & 61.57 & & 77.14\\
    CNN-LSTM~\cite{guera2018deepfake} & & 67.91 & 63.80 & 50.13 & & 60.61\\
    Freq-SCL~\cite{DBLP:conf/cvpr/LiXLW021} & & 79.09 & 74.21 & 53.99 & & 73.53\\
    MultiAtt~\cite{zhao2021multi} & & 74.56 & 80.61 & 60.90 & & 72.02\\
    LAST~(ours) & & {\textbf{89.60}} & {80.66} & {\textbf{67.44}} & & \textbf{79.23} \\ %94.21
    %\bottomrule
    \hhline{|-------|}
  \end{tabular}
  }
  \vspace{-1em}
\end{table}

\begin{table*}[ht]
%\tiny
\small
\caption{Robustness evaluation when trained on uncompressed FF++ and evaluated under seven different perturbations described in~\cite{DBLP:conf/cvpr/JiangLW0L20}~(video-level AUC~(\%)).}\label{tab:robust}
   \centering
   \setlength{\tabcolsep}{3.5mm}{
   %\resizebox{\linewidth}{!}{
       \begin{tabular}{lccccccccl}
          %\toprule
          \hhline{|----------|}
              {Method} &
              \textcolor{gray}{Clean} & {Saturation} & {Contrast} & {Block} & {Noise} & {Blur} & {Pixel} & {Compress} & \textit{Avg/Drop}\\
              \hhline{|----------|}
              {Xception~\cite{Rssler2019FaceForensics}} & \textcolor{gray}{99.8} & 
              99.3 & 98.6 & \textbf{99.7} & 53.8 & 60.2 & 74.2 & 62.1 & 78.3/-21.5\\
              {CNN-aug~\cite{DBLP:conf/cvpr/WangW0OE20}} & \textcolor{gray}{99.8} & 
              99.3 & 99.1 & 95.2 & 54.7 & 76.5 & 91.2 & 72.5 & 84.1/-15.7\\
              {Patch-based~\cite{DBLP:conf/eccv/ChaiBLI20}} & \textcolor{gray}{99.9} & 
              84.3 & 74.2 & 99.2 & 50.0 & 54.4 & 56.7 & 53.4 & 67.5/-32.4\\
              
              {X-Ray~\cite{li2020face}} & \textcolor{gray}{99.8} & 
              97.6 & 88.5 & 99.1 & 49.8 & 63.8 & 88.6 & 55.2 & 77.5/-22.3\\
              {CNN-GRU~\cite{DBLP:conf/cvpr/SabirCJAMN19}} & \textcolor{gray}{99.9} & 
              99.0 & 98.8 & 97.9 & 47.9 & 71.5 & 86.5 & 74.5 & 82.3/-17.6\\
              {LipForensics} \cite{DBLP:conf/cvpr/HaliassosVPP21} & \textcolor{gray}{99.9} & \textbf{99.9} & {99.6} & 87.4 & 73.8 & 96.1 & 95.6 & 95.6 & 92.6/-7.3\\
              {FTCN} \cite{DBLP:conf/iccv/ZhengB0ZW21} & \textcolor{gray}{99.4} & 
              99.4 & 96.7 & 97.1 & 53.1 & 95.8 & {98.2} & 86.4 & 89.5/-9.9\\
              LTTD~\cite{DBLP:conf/nips/GuanZHD0QZ22} & \textcolor{gray}{99.4} & 
              98.9 & 96.4 & 96.1 & 82.6 & 97.5 & {98.6} & 95.0 & 95.0/-4.3\\
              RealForensics~\cite{DBLP:conf/cvpr/HaliassosMPP22} & \textcolor{gray}{99.8} & 99.8 & {99.6} & 98.9 & 79.7 & 95.3 & {98.4} & 97.6 & 95.6/-4.2\\
              AltFreezing~\cite{wang2023altfreezing} & \textcolor{gray}{99.9} & 99.5 & \textbf{99.8} & 97.1 & 75.2 & 97.4 & 98.1 & 92.6 & 94.2/-5.7\\
              {LAST~(ours)} & \textcolor{gray}{99.9} & 
              {99.6} & 99.2 & {99.1} & \textbf{87.8} & \textbf{98.2} & \textbf{98.8} & \textbf{97.8} & \textbf{97.2}/-\textbf{2.7}\\
          %\bottomrule
          \hhline{|----------|}
       \end{tabular}
   }
   %\vspace{-.75em}
\end{table*}

\subsection{Experimental Results}
\label{sec:exp_res}

\myPara{Cross-dataset generalization.}
%The detector's generalization to unknown distributions is a major challenge.
To evaluate the generalization of our adaptation paradigm, we conduct cross-dataset experiments, which means the training and testing videos are from different datasets' distributions, and we regard the testing videos as unlabeled target videos. Specifically, we train our model on FF++ and evaluate on other three datasets and the results are presented in Tab.~\ref{tab:cross-dataset}.
We observe that our method consistently outperforms other recent detectors on cross-dataset generalization, such as average 11.85\% and 6.02\% AUC improvement compared to recent NiCL~\cite{qiao2024deepfake} and ID$_3$~\cite{yin2024improving}.
%such as achieving 74.88\%, 86.43\%, and 92.01\% AUC on DFDC, Celeb-DF, and DFD respectively. 
This provides more support for the generalization of our proposed LAST by adapting to the spatiotemporal clues of unknown target videos in latent space via recovering them.

Furthermore, to support the flexibility of our method, we make comparisons with other detectors on architectures and trainable parameters for usual face forgery detection task, as illustrated in Tab.~\ref{tab:params-arch}, where our detector has only 4.5M parameters of linear layers to optimize for adaptation, supporting its high efficiency and low-cost.

\myPara{Cross-manipulation generalization.}
We further evaluate the cross-manipulation generalization, which means the real source videos are from the same distribution but employ different forgery methods. Since the FF++ dataset includes four different subset generated by different forgery methods, we choose one of them as the source labeled domain and other three as the unlabeled target domains. The results are shown in Tab.~\ref{tab:cross-manipulation}. 
We observe that our method still achieves impressive generalization when detecting unseen methods, with average 11.44\% and 5.17\% AUC improvements compared to Xception~\cite{Rssler2019FaceForensics} and MultiAtt~\cite{zhao2021multi}. 
Both the cross-dataset and cross-manipulation results provide significant evidence for the impressive generalization of our proposed LAST when detecting unseen datasets and methods by adapting to the spatiotemporal clues in a semi-supervised manner.

\myPara{Robustness evaluation.}
%The robustness of detectors to common perturbations is also a challenging issue in practical scenarios, such as the compression in social media.
We further evaluate the robustness of our method under unseen different perturbations by training on uncompressed videos~(raw) but evaluating on videos added perturbations on FF++. We consider the following seven perturbations described in~\cite{DBLP:conf/cvpr/JiangLW0L20}: color saturation, color contrast, block-wise, Gaussian noise, Gaussian blur, pixelation and video compression, where each of them has five different severity levels. 
We report the average AUC scores across five severity levels and make comparisons as presented in Tab.~\ref{tab:robust}.
We observe that our method is more robust than the existing detectors on most of the unseen perturbations, only -2.7\% average AUC drop compared to no perturbation, demonstrating its impressive robustness, especially on noise, blur, pixel, and compression. We analyze the reason is because the distorted spatiotemporal clues of the target videos in above situations are also fully leveraged by our adaptation paradigm to minimize the distribution gap.
The results provide evidence for the impressive robustness of our method on most common unseen perturbations in real detecting scenarios by adapting the spatiotemporal clues in a semi-supervised manner.

\subsection{Ablation Study}
\label{sec:ablation}
We conduct further ablation studies to analyze the adaptation paradigm, and the initialization stage under both intra- and cross-dataset settings. Unless stated otherwise, we train and test on FF++ for intra-dataet settings, and train on FF++ while test on Celeb-DF for cross-dataset settings.

\begin{table}[ht]
\small
\caption{Analysis of the proposed LAST paradigm, where we train on one dataset and evaluate on the other two.}\label{tab:ablation-cross-three}
   \centering
   \resizebox{\linewidth}{!}{
   \begin{tabular}{c|c|cccc}
      %\toprule
      \hhline{|------|}
          {finetune Set} & {Test Set} & {{LAST}} & {ACC(\%)$\uparrow$} & {AUC(\%)$\uparrow$} & {EER(\%)$\downarrow$} \\
          \hhline{|------|}
          \multirow{4}{*}{FF++}
          & \multirow{2}{*}{DFDC} & w/o & 84.25 & 72.97 & 34.05 \\ %\hhline{~~|----|}
          & & w & 85.28 & 74.88 & 31.45\\
          \hhline{~|-----|}
          & \multirow{2}{*}{Celeb-DF} & w/o & 77.03 & 79.82 & 26.76\\
          & & w & 81.66 & 86.43 & 20.59\\
          %\cmidrule(lr){2-6}
          \hhline{|------|}
          %\midrule
          \multirow{4}{*}{DFDC}
          & \multirow{2}{*}{FF++} & w/o & 70.05 & 72.82 & 31.18 \\
          & & w & 79.19 & 87.02 & 20.43\\
          \hhline{~|-----|}
          & \multirow{2}{*}{Celeb-DF} & w/o & 68.34 & 66.20 & 36.47\\
          & & w & 73.75 & 74.05 & 35.01\\
          %\cmidrule(lr){2-6}
          \hhline{|------|}
          \multirow{4}{*}{Celeb-DF}
          & \multirow{2}{*}{FF++} & w/o & 85.28 & 90.74 & 15.05 \\
          & & w & 87.82 & 94.95 & 15.04\\
          \hhline{~|-----|}
          & \multirow{2}{*}{DFDC} & w/o & 84.25 & 66.44 & 38.93\\
          & & w & 84.51 & 73.40 & 34.05\\
      \hhline{|------|}
   \end{tabular}
   }
   %\vspace{-.75em}
\end{table}

\myPara{Latent spatiotemporal adaptation.}
We first analyze the effectiveness and generalization of our latent spatiotemporal adaptation paradigm by training on one and evaluating on other two datasets with/without using the adaptation. Not using the adaptation means we directly perform the supervised training on labeled source videos then testing on unlabeled target videos.
The results are shown in Tab.~\ref{tab:ablation-cross-three}.
We observe that our proposed paradigm can achieve an impressive generalization improvement under cross-dataset setting, with an average 3.83\% improvement on ACC, 6.96\% improvement on AUC, and 4.31\% improvement on EER compared to not employed.
The results provide more evidence of the improvement in generalization achieved by both our proposed LAST paradigm and further indicate that our method is flexible to properly handle various cross-dataset situations. 

\begin{figure}[htb]
   \centering
   \includegraphics[width=\linewidth]{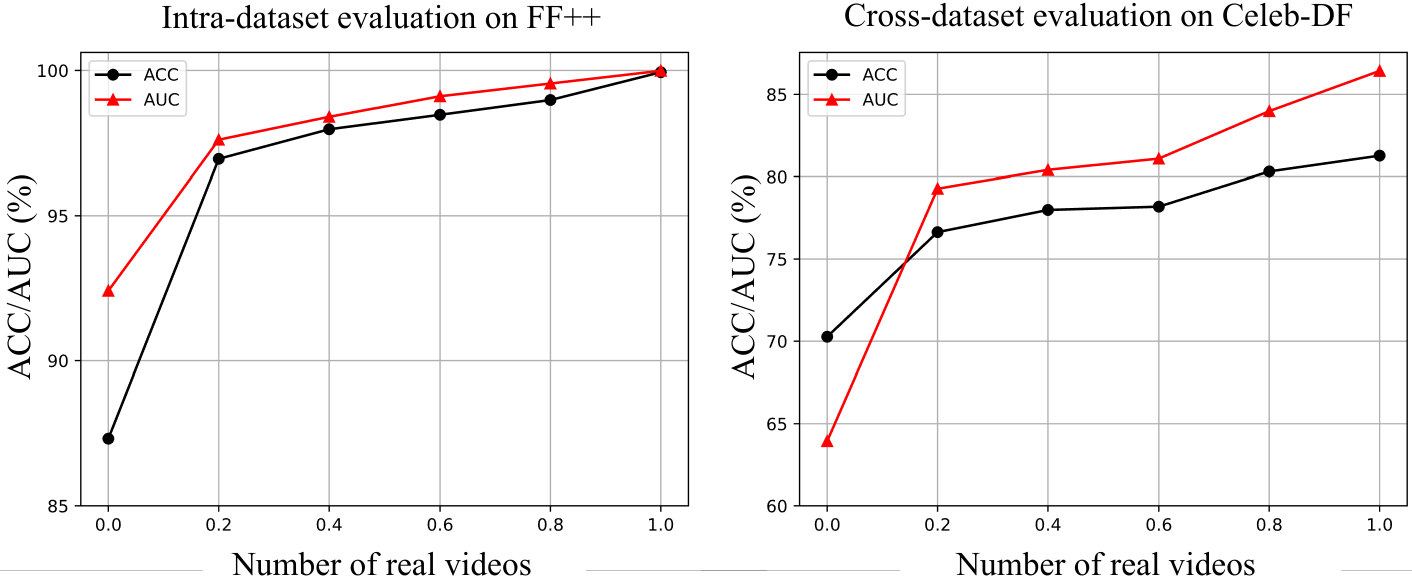}
   \caption{\label{figure:data-scale} Effect of the common spatiotemporal initialization under both intra- and cross-dataset settings.}
   \vspace{-.5em}
\end{figure}

\myPara{Common spatiotemporal initialization.}
We then investigate the impact of the common spatiotemporal initialization on real-only videos. Specifically, we explore the effect by using different amounts of real videos for initialization.
We set the data scale from 0.0 to 1.0 with an interval of 0.2 and use the ACC and AUC to evaluate the performance, noticing that setting the data scale to 0.0 means there is no initialization stage for common spatiotemporal representation learning and we directly employ adaptation with random initialization. The results are presented in Fig.~\ref{figure:data-scale}.
We observe that the ACC and AUC in both settings increase with the data scale increasing, and the increase is more significant on cross-dataset evaluation, suggesting the effectiveness of our proposed initialization method on real videos to eliminate the influence of specific forgery videos and achieve improved generalization.

\begin{figure}[b]
   \centering
   \vspace{-1em}
   \includegraphics[width=.75\linewidth]{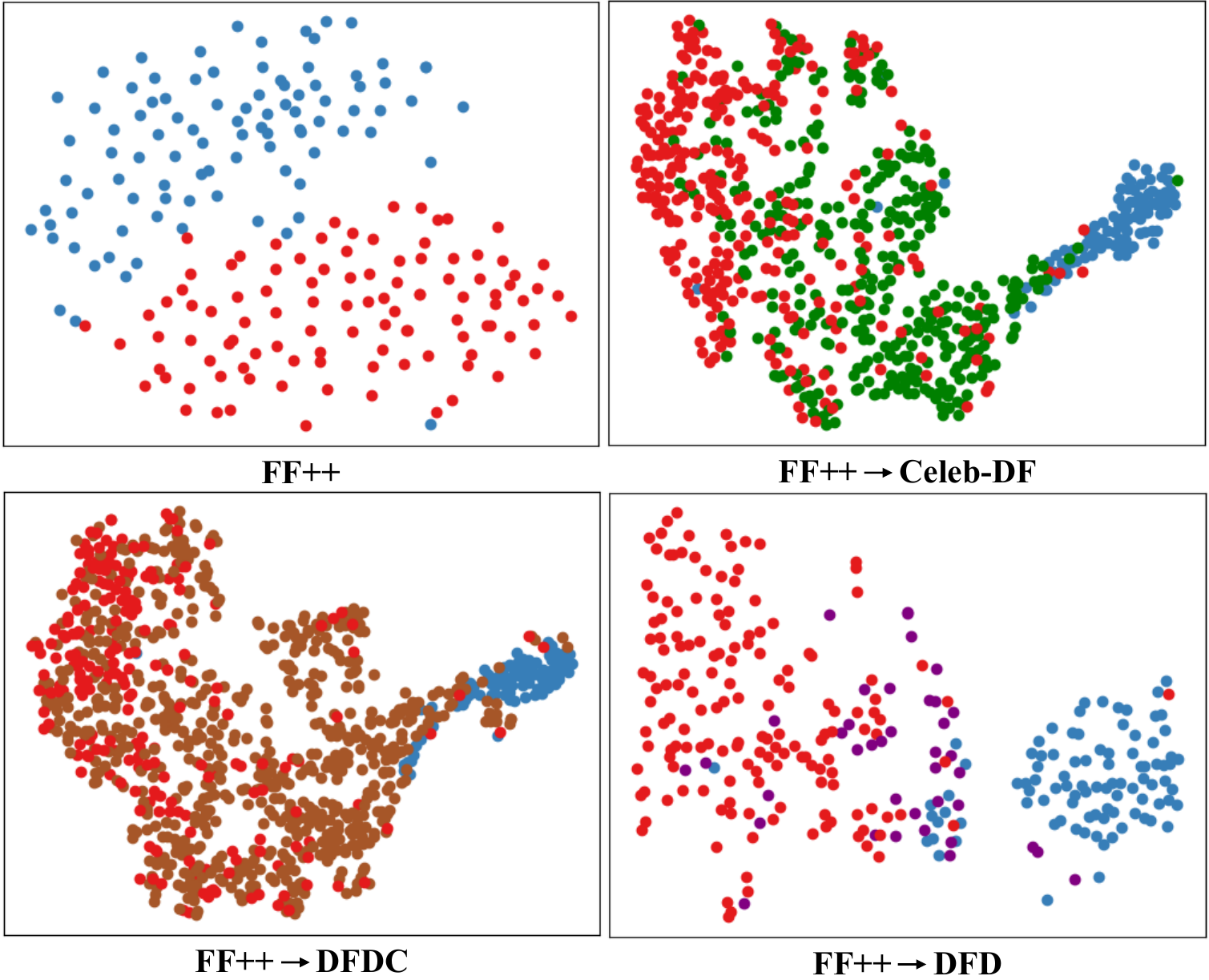}
   \caption{\label{figure:tsne} The t-SNE~\cite{tSNE} visualization of the spatiotemporal representation under both intra-~(top left) and cross-dataset~(other three) settings. 
   }
   %\vspace{-2mm}
\end{figure}

\subsection{Visualization Analysis}
\myPara{Representation.}
To analyze whether our proposed method could effectively bridge the gaps between unknown and unknown videos, we visualize the spatiotemporal representation learned by our LAST in both intra- and cross-dataset settings using t-SNE~\cite{tSNE}. 
Specifically, we visualize the feature learned from the adaptive layer~$L_d(\cdot)$ as the representation being adapted.
The results are shown in Fig.~\ref{figure:tsne}. From the results, we first observe that the learned representations of real and fake videos are clustered with a clear discrepancy margin in latent space for intra-dataset evaluation, which indicates that our model still preserve strong discriminability to distinguish forged videos from real ones. 
Besides, the fake videos from different datasets are also closely clustered and distinguishable from real videos in latent space, which suggests that our proposed LAST paradigm could effectively bridge the latent gap of different forgery videos to achieve improved generalization.

\begin{figure}[ht]
   \centering
   \includegraphics[width=\linewidth]{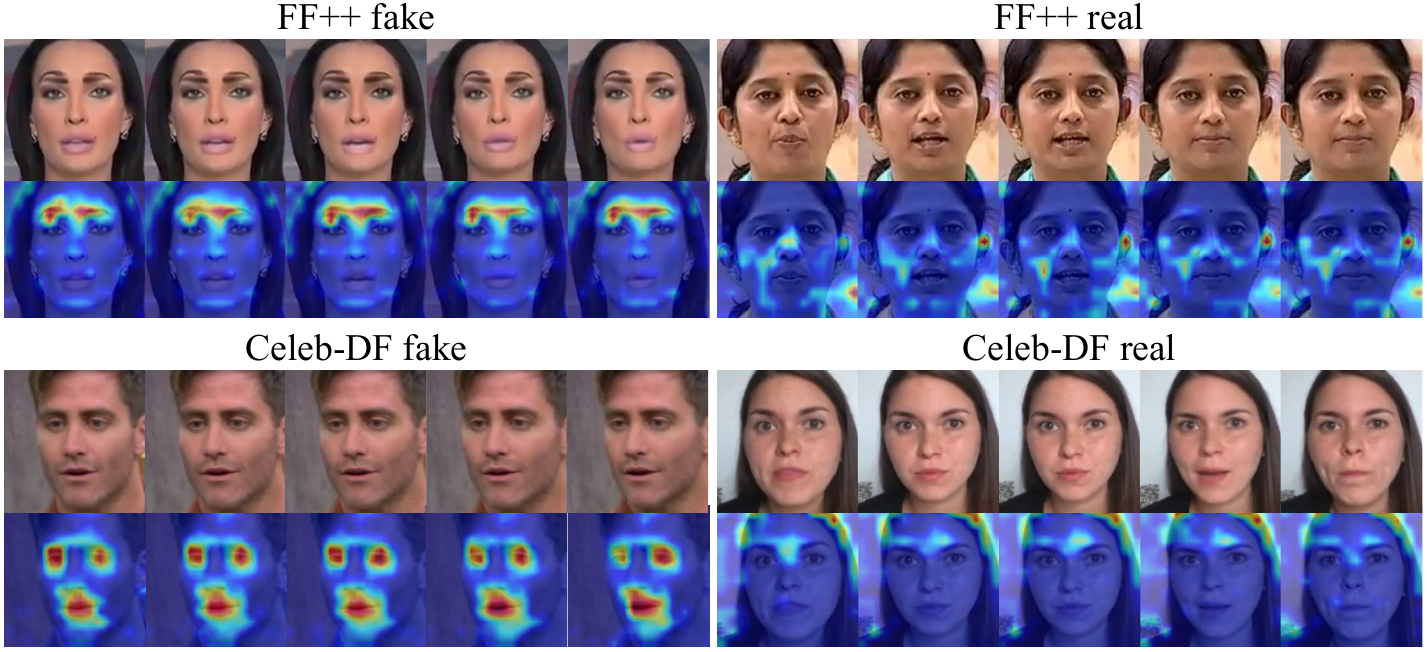}
   \caption{\label{figure:grad-cam} Grad-CAM~\cite{DBLP:conf/iccv/SelvarajuCDVPB17} results under both intra- and cross-dataset settings. We find that our proposed method can effectively respond to the forgery traces.}
   %\vspace{-.75em}
\end{figure}

\myPara{Interpretability.}
To evaluate whether our proposed method can respond to the spatiotemporal clues, we provide the Grad-CAM~\cite{DBLP:conf/iccv/SelvarajuCDVPB17} heatmaps based on the recovered local spatial features of each frame under both intra- and cross-dataset settings as shown in Fig.~\ref{figure:grad-cam}. We observe that the heatmap of the forgery data focuses on the specific forgery traces, such as the forehead face boundary in FF++ and the central face area, including eyes and mouth in Celeb-DF. The difference is highly related to the different spatiotemporal clues left by different forgery methods, which is consistent with the intuition of our proposed method. The heatmap for the real data is average on the whole facial area since there are no forgery traces. The results demonstrate the effectiveness of our method from the decision-making perspective and provide human-understandable explanations for detection results.
%Please refer to our supplementary material for more examples.

\section{Conclusion}
\label{sec:conclu}
In this paper, we propose a Latent Spatiotemporal Adaptation~(LAST) approach to achieve generalized face forgery video detection. We first model the spatiotemporal latent space by incorporating a CNN for local spatial features of single frame then cascaded by a vision transformer for long-term spatiotemporal representations. Further, by optimizing a linear head to perform the usual forgery detection task and recover the spatial clues from learned spatiotemporal representation in a semi-supervised manner, our method could flexibly adapt to unknown target videos' spatiotemporal patterns, leading to improved generalization. Additionally, to eliminate the effect of specific forgery videos in the initial latent space, we propose to pre-train our CNN and transformer on real-only face videos to learn the common spatiotemporal representation of face videos with two simple yet effective self-supervised tasks: reconstruction and contrastive learning, and further keep them frozen during adaptation. Extensive experimental results demonstrate the superiority of our method over state-of-the-art competitors with impressive generalization and robustness.
In the future, we aim to apply our method to more complicated cases in real-scenarios and also extend our method to other media forensic tasks, such as audio and multi-modalities.

\clearpage

\bibliographystyle{IEEEtran}
\bibliography{IEEEabrv, main}

\vfill

\end{document}